\documentclass[10pt]{article} 
\usepackage[accepted]{rlc}

\usepackage{amssymb}            
\usepackage{mathtools}          
\usepackage{mathrsfs}           
\mathtoolsset{showonlyrefs}     
\usepackage{graphicx}           
\usepackage{subcaption}         
\usepackage[space]{grffile}     
\usepackage{url}    
\usepackage{amsmath}

\usepackage{wrapfig}

\usepackage{algorithm}

\let\classAND\AND
\let\AND\relax
\usepackage{algorithmic}

\let\AND\classAND
\AtBeginEnvironment{algorithmic}{\let\AND\algoAND}

\title{Investigating the Interplay of Prioritized Replay and Generalization}

\author{Parham Mohammad Panahi$^1$, Andrew Patterson$^1$, Martha White$^{1,2,3}$, Adam White$^{1,2,3}$\\
$^{1}$ University of Alberta \ \
$^{2}$ Alberta Machine Intelligence Institute (Amii) \ \
$^{3}$ CIFAR AI Chair\\
$\{$parham1, ap3, whitem, amw8$\}$@ualberta.ca
}

\begin{document}

\maketitle

\begin{abstract}

Experience replay, the reuse of past data to improve sample efficiency, is ubiquitous in reinforcement learning. Though a variety of smart sampling schemes have been introduced to improve performance, uniform sampling by far remains the most common approach. One exception is Prioritized Experience Replay (PER), where sampling is done proportionally to TD errors, inspired by the success of prioritized sweeping in dynamic programming. The original work on PER showed improvements in Atari, but follow-up results were mixed. In this paper, we investigate several variations on PER, to attempt to understand where and when PER may be useful. Our findings in prediction tasks reveal that while PER can improve value propagation in tabular settings, behavior is significantly different when combined with neural networks. Certain mitigations---like delaying target network updates to control generalization and using estimates of expected TD errors in PER to avoid chasing stochasticity---can avoid large spikes in error with PER and neural networks but generally do not outperform uniform replay. In control tasks, none of the prioritized variants consistently outperform uniform replay. We present new insight into the interaction between prioritization, bootstrapping, and neural networks and propose several improvements for PER in tabular settings and noisy domains.

\end{abstract}

\section{Introduction}

Experience Replay (ER) is widely used in deep reinforcement learning (RL) and appears critical for good performance. The core idea of ER is to record transitions (experiences) in a memory, called a buffer, replay them by sub-sampling mini-batches to update the agent's value function and policy. ER allows great flexibility in agent design. ER can be used to learn from human demonstrations (pre-filling the replay buffer with human data) allowing off-line pre-training and fine-tuning. ER has been used to learn many value functions in parallel, as in Hindsight ER \citep{andrychowicz2018hindsight}, Universal Value Function Approximators \citep{schaul2015universal}, and Auxiliary Task Learning \citep{jaderberg2016reinforcement,wang2024investigating}. ER can be seen as a form of model-based RL where the replay buffer acts as a non-parametric model of the world \citep{pan2018organizing,van2019use}, or ER can be used to directly improve model-based RL systems \citep{lu2024synthetic}. In addition, ER can be used to mitigate forgetting in continual learning systems~\citep{anand2024prediction}. ER has proven crucial for mitigating the sample efficiency challenges of online RL, as well as mitigating instability due to off-policy updates and non-stationary bootstrap targets. The most popular alternative, asynchronous training, requires multiple copies of the environment, which is not feasible in all domains and typically makes use of a buffer anyway (e.g., \cite{horgan2018distributed}).

There are many different ways ER can be implemented. The most widely used variant, i.i.d or uniform replay, samples experiences from the buffer with equal probability. As discussed in the original paper \citep{lin1991programming}, ER can be combined with lambda-returns and various sampling methods. Experience can be sampled in reverse order it occurred, starting at terminal states. Transitions can be sampled from a priority queue ordered by TD errors---the idea being transitions that caused large updates are more important and should be resampled. Samples can be drawn with or without replacement---avoiding saturating the mini-batch with high priority transitions.  The priorities can be periodically updated. We could use importance sampling to re-weight the distribution in the queue, and generally we could dynamically change the distribution during the course of learning. Despite the multitude of possible variants~\citep{igata2021prioritized,sun2020attentive,kumar2023introspective,lee2019sampleefficient,hong2023topological,kobayashi2024revisiting,li2021revisiting,li2022clustering} simple i.i.d replay remains the most widely used approach.\footnote{See \cite{wittkuhn2021replay} for a nice review.} 
    
The exception to this is Prioritized Experience Replay (PER)~\citep{schaul2016prioritized}, where experience is sampled from the buffer based on TD errors. Like prioritized sweeping that inspired it \citep{moore1993prioritized}, PER in principle should be more efficient than i.i.d sampling. Imagine, a sparse reward task where non-zero reward is only observed at the end of long trajectories; sampling based on TD errors should focus value updates near the terminal state efficiently propagating reward information across the state space. This approach was shown to improve over i.i.d sampling in Atari when combined with Double DQN \citep{schaul2016prioritized}. The results in follow up studies, however, were mixed and did not show a clear benefit for using PER generally \citep{fedus2020revisiting,hessel2017rainbow,li2021revisiting,ma2023revisiting,horgan2018distributed,fu2022benchmarking}. Compared with i.i.d sampling, PER introduces six hyper-parameters controlling importance sampling and how additional experiences are mixed with the prioritized distribution.

In this paper, we explore several different variations of PER in carefully designed experiments in the hopes of better understanding where and when PER is useful. Canonical PER \citep{schaul2016prioritized} uses several additional components that make the impacts of prioritization harder to analyze. We compare PER with several simplified variants using simple chain tasks where value propagation and prioritization should be critical for performance. We find that only in tabular prediction, do all prioritized variants outperform i.i.d replay. Combining basic prioritization with sampling without replacement and updating the priorities in the buffer (things not done in public implementations), further improves performance in the tabular case. Our results show that prioritization, bootstrapping, and function approximation cause problematic over-generalization, possibly motivating the design choices of PER which ultimately causes the method to function more like i.i.d sampling under function approximation. Our results in chain domains with neural network function approximation and across several classic control domains, perhaps unsurprisingly, shows no clear benefit for any prioritized method.  

We also introduce and investigate a natural extension to PER based on ideas from Gradient TD methods \citep{sutton2009fast,patterson2022generalized}. These methods stabilize off-policy TD updates by learning an estimate of the expected TD error. This estimate can be used to compute priorities and is less noisy than using instantaneous TD errors. This expected PER algorithm works well in tabular prediction tasks and noisy counter-examples where PER fails, but is generally worse than i.i.d sampling under function approximation and ties i.i.d in classic control problems---though it appears more stable. Although somewhat of a negative result, expected PERs performance suggests noise is not the explanation for i.i.d sampling's superiority over PER and more research is needed to find generally useful prioritization mechanisms.

\section{Background, Problem Formulation, and Notation}
In this paper, we investigate problems formulated as discrete-time, finite Markov Decision Processes (MDP). On time step, $t$, the agent selects an action $A_t\in\mathcal{A}$ in part based on the current state, $S_t\in\mathcal{S}$. The MDP transitions to a new state $S_{t+1}$ and emits a reward signal $R_{t+1}\in\mathcal{R}$. The agent's action choices are determined by it's policy $A_t\sim\pi(\cdot|S_t)$, and the goal of learning is to adjust $\pi$ to maximize the future expected return $\mathbb{E}_\pi[G_t| S_t=s, A_t=a]$ $=\mathbb{E}_\pi[R_{t+1} + \gamma R_{t+2}+\gamma^2 R_{t+3}+\hdots | S_t=s, A_t=a]$, 
where $\gamma\in[0,1]$. The expectation is dependent on future actions determined by $\pi$ and future states and rewards according to the MDP.

We focus on action-value methods for learning $\pi$. 
In particular, Q-learning estimates the state-action value function $q_\pi(s,a)\doteq\mathbb{E}_\pi[G_t| S_t=s, A_t=a]$
$~\forall~s,a\in\mathcal{S}\times\mathcal{A}$ 
via temporal difference updates from sample interactions: 
$\hat{q}(S_t,A_t)=\hat{q}(S_t,A_t)+\alpha\delta_t$, 
where $\delta_t \doteq R_{t+1} + \gamma \max_a\hat{q}(S_{t+1},\cdot) - \hat{q}(S_t,A_t)$ is called the TD-error with learning-rate parameter $\alpha\in\mathbb{R}^+$.
Actions are selected according to an $\epsilon$-greedy policy: 
selecting $A_t= \arg\max \hat{q}(S_t,\cdot)$
~$1-\epsilon$ percentage of the time and a random action otherwise. 
In many tasks, it is not feasible to learn an action-value for every state. In these cases, we use a non-linear parametric approximation of the value, $\hat{q}_{\bf w}(s,a)\approx q_\pi(s,a)$, where ${\bf w}$ are the parameters of a neural network (NN), which are adjusted via semi-gradient Q-learning rule.

Semi-gradient Q-learning when combined with NNs is often unstable, and so DQN is often preferred. The DQN algorithm combines: target networks, ER, and an optimizer \citep{mnih2015human}. Target Networks replace $\max_a\hat{q}_{\bf w}(S_{t+1},\cdot)$ in the TD-error with an older copy of the network. In this paper we use the Adam optimizer \citep{Kingma2015adam}. Experience Replay is used to perform mini-batch updates to $\hat{q}_{\bf w}$ from a finite, first-in-first-out buffer. Sampling from the buffer is uniform or i.i.d meaning the value estimate on the current step is updated based experiences observed in the recent past, not necessarily the most recent transition. The appendix contains the pseudo code for DQN and its key hyperparameters.

In control tasks we learn $\hat{q}$ and $\pi$, however, in prediction tasks $\pi$ is given and fixed and we are interested in learning the state-value function: $\hat{v}(s) \approx v_\pi(s)\doteq\mathbb{E}_\pi[G_t| S_t=s]$. This can be done using the Temporal Difference learning algorithm~\citep{sutton1988learning}, the state-value analog of TD update above, which has a semi-gradient variant for learning $\hat{v}_{\bf w}: \mathcal{S}\rightarrow\mathbb{R}$. See \cite{sutton2018reinforcement} for an extensive overview of all these topics.

\def\pername{Naive PER}

\section{Variants of Prioritized Replay}
In this section, we define two variants of PER that we use to better understand the role of prioritization. We use the name DM-PER for \cite{schaul2016prioritized}'s prioritized replay algorithm and {\em Uniform} to refer to classic i.i.d replay.   

\begin{wrapfigure}[14]{l}{0.4\columnwidth}
\vspace{-0.75cm}
\begin{center}
\includegraphics[width=0.4\columnwidth]{"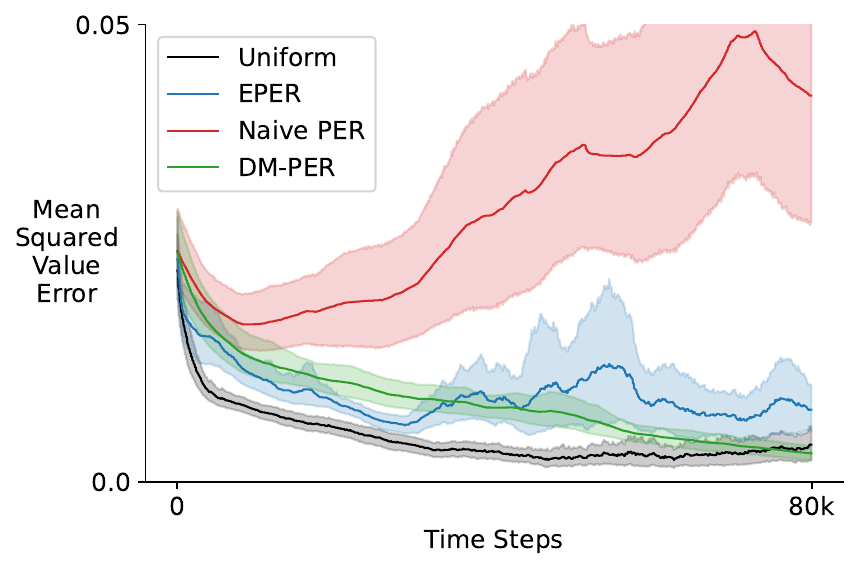"}
\end{center}
\vspace{-0.5cm}
\caption{Prioritization can be problematic in noisy prediction with NNs. Results averaged over 30 trials; shaded region are 95\% bootstrap Confidence Intervals (CI).}
\vspace{0.85cm}
\label{fig:noisy-pred}

\end{wrapfigure}

We start by describing a simplified prioritized replay algorithm, which we will call \emph{\pername}. Starting from a uniform replay, the \pername\ algorithm modifies only the sampling strategy from uniform to proportional to TD error. We record the TD-error as soon as a sample is added to the buffer, then update that TD-error when a transition is sampled. We do not mix in uniform sampling, we do not squash the priorities with an exponential hyperparameter, and we do not use importance weights. In this way, the \pername\ algorithm closely resembles tabular prioritized sweeping, except we sample probabilistic according to the priorities rather than use a priority queue. The full pseudocode for \pername\ can be found in the appendix.

In order to study the effects of noise on the prioritization strategy, we introduce a new prioritization variant \emph{expected} PER (EPER). Instead of using the sample TD-error, $\delta_t$, which can be noisy when the reward or the transition dynamics are stochastic, EPER uses an estimate of the expected TD-error $\mathbb{E}\left[ \delta_t \;|\; S_t=s \right]$. 
This expectation averages out random effects from action selection, transition dynamics, and the reward signal. In control, the expectation is conditioned on both state and action.

Learning this expectation can be formulated as a simple least-squares regression problem with samples $\delta_t$ as the target, yielding the following online update rule:
$\theta_{t+1} \leftarrow \theta_t + \alpha (\delta_t - h_\theta(S_t)) \nabla_\theta h_\theta(S_t)$,
where $h_\theta$ is a parametric approximation of $\delta_t$ with parameters $\theta$.
This secondary estimator forms the basis of the gradient TD family of methods \citep{sutton2009fast,patterson2022generalized} making it natural to combine with recent gradient TD algorithms such as EQRC \citep{patterson2022generalized}. In other words, if we use EQRC instead of DQN, we can use EPER to attain a less noisy signal for computing priorities with no extra work because EQRC is estimating $h_\theta$ anyway.

Figure \ref{fig:noisy-pred} demonstrates the potential benefits of EPER over \pername. The task is to estimate the state-value function of a random policy in a Markov chain with the only reward on the terminal transition (described in more detail in the next section). The terminal reward is polluted by zero mean non-symmetric noise. The hyperparameters of all methods are systematically tuned, and still we see \pername\ is negatively impacted. The DM-PER algorithm is robust in this case, which is not surprising given its use of importance sampling and mixing in i.i.d samples. EPER is not as robust, but achieves this with a much simpler approach.   

\section{An Empirical Investigation of Prioritization in Replay}

The idea of prioritized replay is based on the tabular notion of value propagation and the interplay between neural network generalization and prioritized replay remains an open question. This section explores the combined effect of prioritized replay and neural network generalization in RL agents.

\subsection{Comparing Sample Efficiency in Prediction}
In this section we ask several questions in a sparse reward task where rapid value propagation should require careful sampling from the replay buffer. Does naive prioritization improve performance over uniform replay? Do the additional tricks in DM-PER reduce the efficiency of value propagation when they are not really required? Finally, does robustness to noisy TD errors, as in EPER, matter in practice? We investigate these questions with tabular and neural network representations. 

\begin{figure*}[h]
\vspace{-0.5cm}
\begin{center}
\includegraphics[width=0.45\columnwidth]{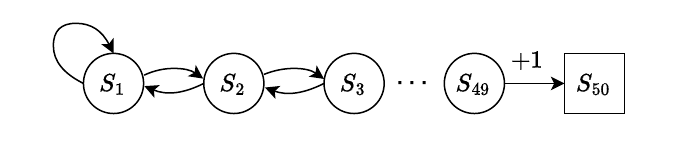}
\vspace{-0.5cm}
\caption{The 50-state Markov chain environment.}
\label{fig:chain-visual}
\end{center}
\vspace{-0.5cm}
\end{figure*}

We consider both policy evaluation and control problems in a 50-state Markov chain environment visualized in Figure \ref{fig:chain-visual}. This is an episodic environment with $\gamma=0.99$ chosen to present a difficult value propagation problem. In every episode of interaction, the agent starts at the leftmost state and at each step takes the \verb+left+ or \verb+right+ action which moves it the corresponding neighbour state. The only reward in this environment is +1 when reaching the rightmost state at which point the episode terminates.

In the policy evaluation experiments, the objective is to estimate the state value function of the random policy. The data for the replay buffer is generated by running the random policy, making this an on-policy prediction task. The performance measure is the Mean Squared Value Error (MSVE) between estimated value function and true value function: $\text{MSVE}({\bf w}) = \sum_s d(s)(v_\pi(s) - \hat{v}_{\bf w}(s))^2$ where $d(s)$ is the state visitation distribution under the uniform policy. In this experiment we have two settings, one where the value function is tabular and one where it is approximated by a two layer neural network with 32 hidden units in each layer and rectified linear unit (ReLU). We systematically tested a broad set of learning rates, buffer size, and batch sizes---over 50 combinations with 30 seeds each. The sensitivity to learning rate can be found in the Appendix. In Figure \ref{fig:chain-pred} we shown a representative result with batch size 8, buffer size 8000, and learning rate $8^{-4}$ in the tabular setting and $8^{-5}$ in the neural network setting. The remaining results are in the Appendix.

Figure \ref{fig:chain-pred} shows the learning curves of different replay methods for policy evaluation in the 50-state Markov chain over time. All three prioritized replay variants perform similarly and they are more sample efficient than uniform replay. The heatmaps show estimated values across states over time. Comparing the heatmap of tabular uniform replay with tabular \pername\ shows an increase in value propagation through the chain when using prioritization.

\begin{figure}[htb]
  \begin{minipage}[c]{0.6\columnwidth}
\includegraphics[width=1.0\columnwidth]{"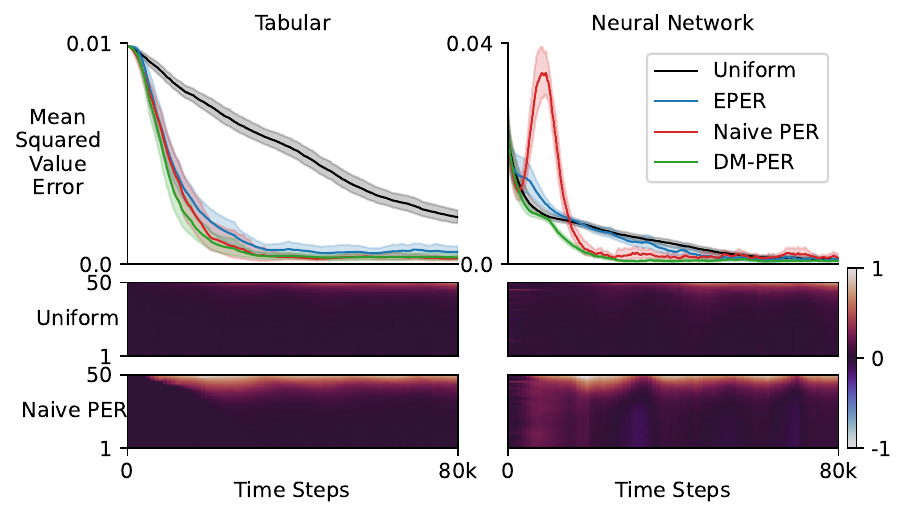"}
  \end{minipage}\hfill
  \begin{minipage}[c]{0.38\columnwidth}
\caption{Prioritized methods can improve sample efficiency in prediction on the 50-state chain in tabular (left) and NN prediction (right). With NN function approximation \pername\ exhibits an increase in MSVE during early learning. The heatmaps show estimated values of the states, 1 to 50, over time. Results are averaged over 30 seeds; shaded regions are 95\% bootstrap CI.}
\label{fig:chain-pred}
  \end{minipage}
\vspace{-.7cm}
\end{figure}

In the neural network setting, the error of \pername\ increases during early learning and then drops to the level of other prioritized replay methods. The gap between other prioritized methods (DM-PER and EPER) and uniform replay is smaller in the neural network setting compared with the tabular setting. Additionally, these two prioritized methods do not exhibit an increase in the MSVE like \pername. 

Perhaps \pername\ over-samples a few transitions which causes the network to spend a lot of its capacity minimizing the error of those transitions at the cost of a worse prediction in other states. It is possible that EPER can mitigate the over-sampling issue because the initial estimates are randomized which helps avoid over-sampling certain transitions. DM-PER reduces the negative effect of over-sampling by using importance sampling weights to reduce the magnitude of updates with high priority transitions.

\begin{figure*}[h]
\vspace{-0.5cm}
\begin{center}
\includegraphics[width=0.9\columnwidth]{"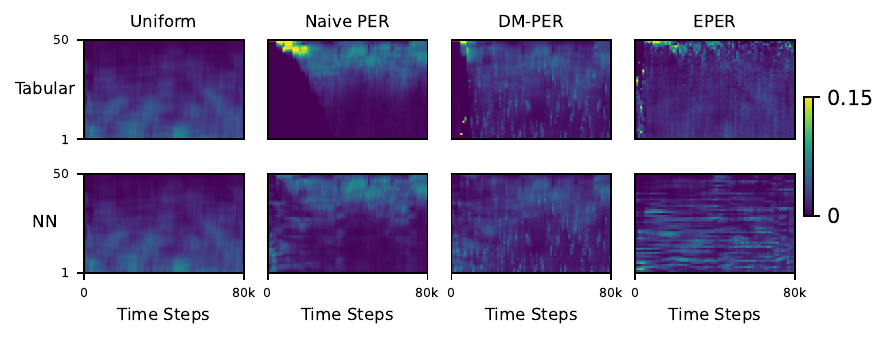"}
\vspace{-0.4cm}
\caption{Probability of sampling a transition starting from each state (1 to 50) from the buffer at each time point, in the 50-state Markov chain for one run.}
\label{fig:sampling-prob}
\end{center}
\vspace{-0.4cm}
\end{figure*}

To better understand what is happening, we visualize the probability of updating a state over time in Figure \ref{fig:sampling-prob}. We calculate this probability by summing over the probabilities of sampling a transition starting from a given state at a given time based on transitions in the replay buffer. The heatmap presents these probabilities, computed every 1000 states over one run. We use the same hyperparameter settings as in Figure \ref{fig:chain-pred}.

The sampling probabilities under uniform replay reflect the state visitation of the random policy, putting a higher probability mass on earlier states.
The sampling distribution of tabular \pername\ follows the intuition from prioritized sweeping by putting most of the probability mass on the rewarding transition at the end of chain, then, increasing the probability of nearby states in a backward fashion to help value propagation. Under neural network function approximation the pattern is similar but more uniform. This is caused by the random initialization of network parameters which generates non-zero TD errors across the state space. The sampling distribution of both DM-PER and EPER are, on the other hand, more structured. Both feature non-terminal transitions with high probability (bright spots) and striping. It is hard to speculate why this occurs, nevertheless, these patterns provide evidence that combining prioritization with NNs can result distributions very different from the tabular case.

\subsection{Overestimation due to Prioritization, Generalization, and Bootstrapping}
In the previous section we saw that \pername\ exhibited a spike in early learning, but why? One possible explanation is that \pername\ is oversampling the terminal transition which causes the NN to inappropriately over-estimate nearby states, causing more oversampling, and so on, spreading across all states.  The heatmap in Figure \ref{fig:sampling-prob} provides some evidence of this. One way to prevent over-generalization is to employ target networks to reduce the effect of bootstrapping.
Note that the task is on-policy prediction, and we do not need a target network to stabilize learning. We use the same setup as in figure \ref{fig:chain-pred} and only consider \pername\ with neural nets.

\begin{figure*}[h]
\vspace{-0.2cm}
\begin{center}
\includegraphics[width=0.9\columnwidth]{"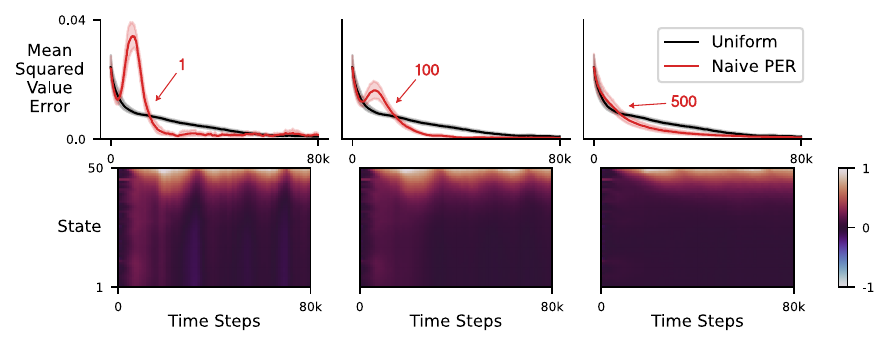"}
\vspace{-0.4cm}
\caption{Target Networks can mitigate \pername's poor performance in the 50-state Markov chain prediction task with NNs. Red numbers above curves indicate Target Network update rate.}
\label{fig:target-net}
\end{center}
\vspace{-0.2cm}
\end{figure*}

The results in Figure \ref{fig:target-net} show the performance of \pername\ with three different target network update rates (1, 100, 500). An update rate of 1 is identical to not using target networks at all.  As we update the target network less frequently we see the spike in the learning curves is reduced. Notice that heatmap for the value function with an update rate of 500 is very similar to \pername\ in the tabular case (see Figure \ref{fig:chain-pred}). We only see a minor performance improvement over uniform replay in Figure \ref{fig:target-net}, but this is expected because updating the target network infrequently is reducing the update rule's ability to propagate value backwards via bootstrapping.

\subsection{Comparing Sample Efficiency in Control}
In this section we turn our attention to a simple control task, again designed in such a way that value propagation via smart sampling should be key. Here our main question is: do the insights about the benefits of prioritization persist when the policy changes and exploration is required.

In the tabular setting, we use Q-learning (without target networks) and in neural network setting we explore two setups: (1) DQN (with target refresh rate of 100) and (2) EQRC (as an alternative method without target network). We report steps to goal as the performance metric for the 50-state Markov chain problem. Buffer size is fixed to 10000, batch size 64, and we pick the best learning rate for each method (see the Appendix for details). Each control agent is run for 100000 steps with an $\epsilon$-greedy policy with $\epsilon = 0.1$.  

\begin{figure}[htb]
  \begin{minipage}[c]{0.75\columnwidth}
\includegraphics[width=1.0\columnwidth]{"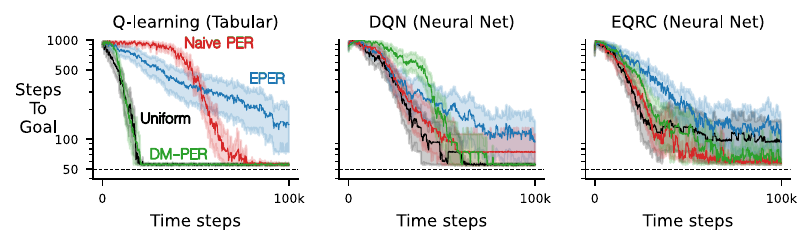"}
  \end{minipage}\hfill
  \begin{minipage}[c]{0.24\columnwidth}
\caption{Prioritization is not more sample efficient than uniform for control in the 50-state Markov chain environment. Results averaged over 50 seeds; shaded regions are 95\% bootstrap CI.}
\label{fig:chain-control}
  \end{minipage}
\vspace{-0.5cm}  
\end{figure}

The results in Figure \ref{fig:chain-control} are somewhat unexpected. The dotted line depicts the performance of the optimal policy. Even in the tabular case, Q-learning with uniform replay is a better than all the three prioritized methods. DM-PER performs just as well as uniform replay, but this could be explained by the fact that DM-PER's sampling is closer to uniform compared with the other prioritization schemes as shown previous in Figure \ref{fig:sampling-prob}. \pername\ eventually reaches the near optimal policy and EPER performs poorly. Under neural network approximation, all tested algorithms have a wide overlapping confidence regions, even with 50 seeds, making differences between methods non conclusive. It seems good performance in prediction does not necessarily translate into improvement in control, even in the same MDP.

\subsection{Sampling Without Replacement \& Updating Priorities}
In this section we explore two simple but natural improvements to replay that could improve performance. There are many possible refinements, and many have been explored in the literature already. Here we select two that have not been deeply explored before, specifically (1) sampling transitions with or without replacement, and (2) recomputing priorities of samples in the buffer. 

When sampling a mini-batch from the replay buffer, one has the option to sample transitions with or without replacement. This decision is important in PER because sampling with replacement can cause a high priority transition to be repeatedly sampled into the same mini-batch. This certainly happens on the first visit to the goal state in the 50 state chain. Uniform replay avoids this problem by design. Most reference implementations of PER sample with replacement. We hypothesize that duplicate transitions in the mini-batch reduces the sample efficiency of prioritized methods, effectively nagating the benefit of mini-batches.

We compare \pername\ with and without replacement sampling and uniform replay in the 50-state Markov chain prediction domain under tabular and neural net function approximation. We used a two layer network with 32 hidden units and ReLU activation, a batch size of 8000 and experimented with several mini-batch sizes (1, 8, 64, 256). With batch size 1, with and without replacement are identical. We report a representative result with learning rate of $8^{-4}$ in the tabular setting and $8^{-5}$ in the neural network setting and report MSVE under the target policy over training time.

\begin{figure*}[h]
\vspace{-0.2cm}
\begin{center}
\includegraphics[width=0.9\columnwidth]{"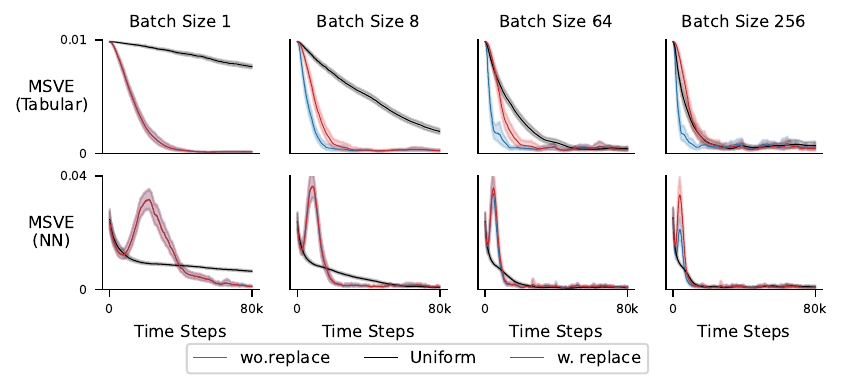"}
\vspace{-0.4cm}
\caption{Sampling without replacement improves the performance of \pername\ in the tabular setting but not with neural nets. Results are averaged over 50 seeds and shaded regions are 95\% bootstrap CI.}
\label{fig:pred-replacement}
\end{center}
\vspace{-0.2cm}
\end{figure*}

Figure \ref{fig:pred-replacement} shows that sampling without replacement provides a minor improvement on the performance of \pername\ in tabular prediction, where \pername\ was already working well, but does not help when combined with NN function approximation. In fact, we again see \pername's characteristic spike due to over-generalization and bootstrapping. This poor performance is somewhat mitigated by larger batch sizes, but still uniform replay is better. Note, as expected, the performance of uniform replay suffers with smaller batch sizes. 

Now we turn to the control setting to evaluate the impact of sampling without replacement. We tested tabular Q-learning and neural network DQN settings. The DQN agent has a two layer network with 32 hidden units and ReLU activation with target refresh rate 100. All agents have buffer size 10000 and a series of batch sizes similar to the previous experiment. The learning rate of each agent is selected by sweeping over a range of step sizes and maximizing over average performance (sweep details in the Appendix). Figure \ref{fig:control-replacement} summarizes the results.

\begin{figure*}[h]
\vspace{-0.2cm}
\begin{center}
\includegraphics[width=0.9\columnwidth]{"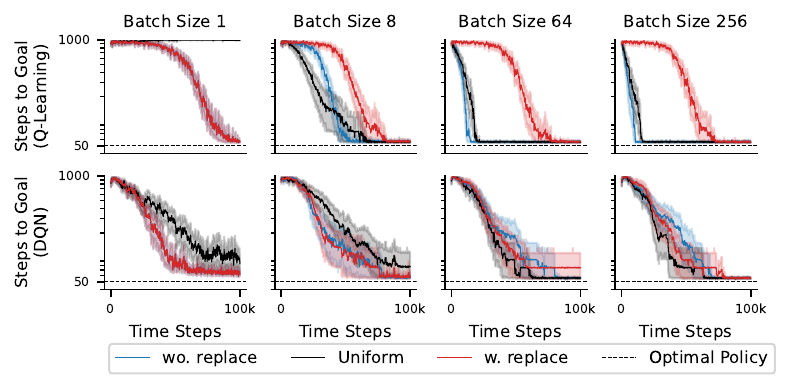"}
\vspace{-0.4cm}
\caption{Sampling with and without replacement in control using \pername\ with tabular and NN representations. Without replacement sampling only helps in the tabular setting. Results averaged over 50 seeds; shaded regions are 95\% bootstrap CI.}
\label{fig:control-replacement}
\end{center}
\vspace{-0.4cm}
\end{figure*}

In tabular control we see a significant improvement in \pername\ when sampling without replacement, whereas with function approximation the result is less clear. In tabular, the gap in performance between with and without replacement steadily increases and eventually \pername\ becomes nearly statistically better than uniform. With DQN (function approximation), larger batch sizes mostly result in ties, though \pername\ without replacement is the only method to always reach optimal on average.  

Taken together, the results which use sampling without replacement suggest a minor benefit. It always helps in the tabular case, at times outperforming uniform replay, and with function approximation it mostly does not hurt performance.

Another factor that can potentially limit the benefit of prioritization is non-informative and outdated priorities in the buffer. The priority of a transition is updated only when the transition is sampled. This means that at any given time the priority of almost all items in the replay buffer are outdated with respect to the current value function. We can update the priority of all transitions in the buffer by recomputing their TD error using the current value function estimate periodically. 

\begin{figure}[htb]
  \begin{minipage}[c]{0.7\columnwidth}
\includegraphics[width=1.0\columnwidth]{"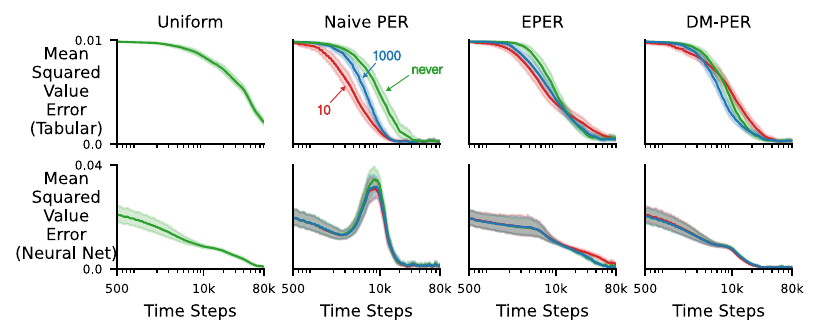"}
  \end{minipage}\hfill
  \begin{minipage}[c]{0.3\columnwidth}
\caption{Recomputing priorities in chain prediction using \pername, EPER, and DM-PER with tabular (top) and NN (bottom) representations. Generally, recomputing does not help. Results averaged over 30 seeds; shaded regions are 95\% bootstrap CI.}
\label{fig:pred-recompute}
  \end{minipage}
\vspace{-0.5cm}  
\end{figure}

We tested this idea in prediction setting in the 50 state chain. We compared the performance of \pername, EPER, and DM-PER recomputing the priorities every 10 and 1000 steps. Again we looked at tabular and NN representations with a two layer neural net of size 32 with ReLU activation for the latter. The buffer size is fixed to 8000, batch size to 8, and learning rate to $8^{-4}$ for tabular and $8^{-5}$ for neural net agents. Figure \ref{fig:pred-recompute} summarizes the results. In short, we see no benefit from recomputing priorities in the function approximation settings and marginal benefit in the tabular case with \pername. Interestingly, for DM-PER recomputing too often, every 1000 steps vs every 10 steps, hurts compared to the default---updating only when a transition is first added or resampled. Note the over-generalization of \pername\ with neural nets is also not reduced more by up-to-date priorities.

\begin{figure}[htb]
  \begin{minipage}[c]{0.7\columnwidth}
\includegraphics[width=1.0\columnwidth]{"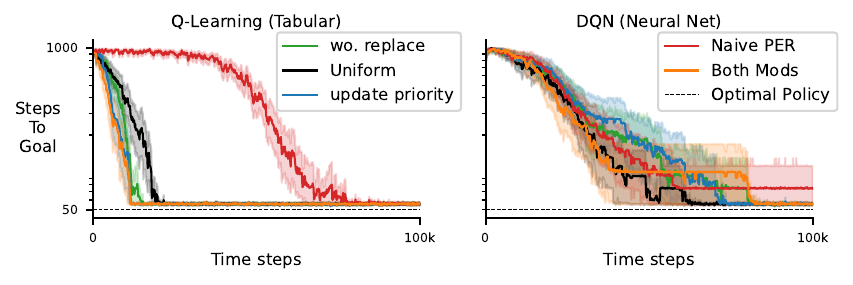"}
  \end{minipage}\hfill
  \begin{minipage}[c]{0.3\columnwidth}
\caption{Combining recomputing priorities and without replacement sampling for tabular (left) and NN (right) control in the chain. Results averaged over 50 seeds; shaded regions are 95\% bootstrap CI.}
\label{fig:control-recompute}
  \end{minipage}
\vspace{-0.5cm}  
\end{figure}

As a final experiment in the chain problem, we investigate if combining sampling without replacement and recomputing priorities every 10 steps, together, can improve the performance of \pername. We conduct this experiment in the control chain problem and repeat the experiment for tabular Q-learning and DQN with a two layer neural net with 32 hidden units and ReLU activation with target refresh rate of 100. The buffer size is fixed to 10000 and batch size is 64, we select the learning rate over a range of values that attained the best average performance. As we see in Figure \ref{fig:control-recompute}, in the tabular case, \pername\ with both modifications achieves the best performance, but barely more than either modification in isolation. Unfortunately, as expected, there is no clear benefit in the function approximation setting. 

\subsection{Comparing Sample Efficiency in Classic Control Domains}
Our previous experiments in the chain were designed to represent an idealized problem to highlight the benefits of smarter replay, and in this section we consider slightly more complex, less ideal tasks. In the chain, we only saw clear advantages for prioritization in prediction and also in control with small batch sizes. The main question here is do these benefits persist or perhaps prioritization will be worse supporting the common preference for uniform replay in deep RL.

We consider four episodic environments which are significantly more complex than the chain, but are small enough that smaller NNs can be used and extensive experimentation is possible. The first three environments, often refered to as classic control feature low dimensional continuous state and discrete actions. MountainCar \citep{moore1990efficient} and Acrobot \citep{sutton1995generalization} are two control tasks where the goal is to manipulate a physical system to get to a goal at the end of a long trajectory. We also include Cartpole due to the unstable dynamics of the balanced position \citep{barto1983neuronlike}. Finally we include the tabular Cliffworld \citep{sutton2018reinforcement} because the reward for falling off the cliff is a large negative value which causes rare but large spikes in the TD error, which might showcase the benefit of EPER. The details about these environments can be found in the Appendix. We set the discount factor $\gamma = 0.99$. The episodes in MountainCar, Acrobot, and Cartpole are cutoff every 500 steps, but there is no episode cutoff in Cliffworld.

\begin{figure}[htb]
  \begin{minipage}[c]{0.75\columnwidth}
\includegraphics[width=1.0\columnwidth]{"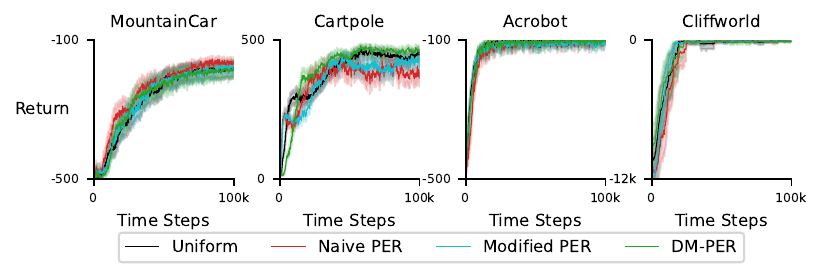"}
  \end{minipage}\hfill
  \begin{minipage}[c]{0.24\columnwidth}
\caption{Performance of DQN replay agents on classic control problems. No clear benefit for prioritization. Results averaged over 100 seeds; shaded regions are 95\% bootstrap CI.}
\label{fig:classic-control}
  \end{minipage}
\vspace{-0.5cm}  
\end{figure}

In this experiment we use DQN with a two layer network of size 64 with ReLU activation and target refresh rate 128. Batch size and buffer size are fixed to 64 and 10000 respectively and the learning rate is selected using a two stage approach to avoid maximization bias \citep{patterson2023empirical}. First each agent is run for 30 seeds sweeping over many learning rate parameter settings, then the hyperparameter which achieved the best average performance is run for 100 new seeds (see details in the Appendix). We include {\em Modified PER}, which combines \pername\ with without-replacement sampling and recomputes the priorities every 10 steps. Figure \ref{fig:classic-control} summarizes the results. Unsurprisingly, prioritization does not improve the sample efficiency over uniform replay in any of the four domains.

\begin{figure}[htb]
  \begin{minipage}[c]{0.75\columnwidth}
\includegraphics[width=1.0\columnwidth]{"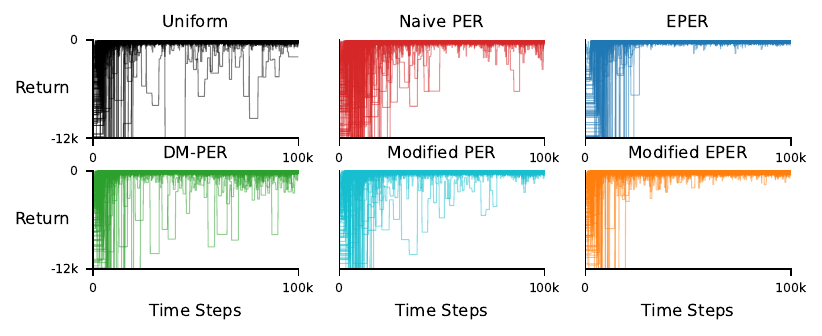"}
  \end{minipage}\hfill
  \begin{minipage}[c]{0.24\columnwidth}
\caption{Performance of 100 individual runs in the Cliffworld shows performance dips using uniform replay, \pername, and DM-PER. EPER-based methods appear to have more stable performance.}\label{fig:cliffworld-seeds}
  \end{minipage}
\vspace{-0.5cm}  
\end{figure}

Looking closely at the learning curve for Cliffworld in Figure \ref{fig:classic-control} we see a small blip in the performance with uniform replay. Recall, we suspected that EPER might show benefit in this MDP due to outlier rewards when the agent falls off the cliff. Average learning curves can hide the stucture of individual runs, so we plotted all the runs individually for each method in 
Figure \ref{fig:cliffworld-seeds}. Here we see DQN with uniform replay periodically performs quite poorly, even late in learning. This is true to a less extent for DM-PER, \pername, and Modified PER. Interestingly, \pername\ variants based on EPER appear substantially more stable with less collapses in performance. 

\section{Conclusion}

In this paper, we conducted a series of carefully designed experiments with prioritized replay under tabular and neural network settings. We found that prioritization combined with non-linear generalization can overestimate values during early learning. It appears that a combination of bootstrapping and neural network generalization is the reason behind this overestimation. Furthermore, we showed in a simple chain domain, several variants of PER outperform i.i.d replay in the prediction setting but have poor sample efficiency in control. Unsurprisingly, no variant of PER improves upon i.i.d replay in classic control domains.

We introduced EPER as a simple modification prioritizing transitions according to a learned estimate of the expected error inspired by gradient TD methods. We showed that EPER can be more robust in noisy reward domains and perform more reliably than PER or i.i.d replay in Cliffworld. Finally, we explored two design decisions in PER, recomputing outdated priorities and sampling batches without replacement, discovering that these additions can improve PER in the tabular setting but have little to no effect when using neural networks.

\bibliography{main}

\begin{thebibliography}{35}
\providecommand{\natexlab}[1]{#1}
\providecommand{\url}[1]{\texttt{#1}}
\expandafter\ifx\csname urlstyle\endcsname\relax
  \providecommand{\doi}[1]{doi: #1}\else
  \providecommand{\doi}{doi: \begingroup \urlstyle{rm}\Url}\fi

\bibitem[Anand \& Precup(2024)Anand and Precup]{anand2024prediction}
Nishanth Anand and Doina Precup.
\newblock Prediction and control in continual reinforcement learning.
\newblock \emph{Advances in Neural Information Processing Systems}, 36, 2024.

\bibitem[Andrychowicz et~al.(2018)Andrychowicz, Wolski, Ray, Schneider, Fong,
  Welinder, McGrew, Tobin, Abbeel, and Zaremba]{andrychowicz2018hindsight}
Marcin Andrychowicz, Filip Wolski, Alex Ray, Jonas Schneider, Rachel Fong,
  Peter Welinder, Bob McGrew, Josh Tobin, Pieter Abbeel, and Wojciech Zaremba.
\newblock Hindsight {{Experience Replay}}.
\newblock 2018.

\bibitem[Barto et~al.(1983)Barto, Sutton, and Anderson]{barto1983neuronlike}
Andrew~G. Barto, Richard~S. Sutton, and Charles~W. Anderson.
\newblock Neuronlike adaptive elements that can solve difficult learning
  control problems.
\newblock \emph{IEEE Transactions on Systems, Man, and Cybernetics},
  SMC-13\penalty0 (5):\penalty0 834--846, 1983.

\bibitem[Fedus et~al.(2020)Fedus, Ramachandran, Agarwal, Bengio, Larochelle,
  Rowland, and Dabney]{fedus2020revisiting}
William Fedus, Prajit Ramachandran, Rishabh Agarwal, Yoshua Bengio, Hugo
  Larochelle, Mark Rowland, and Will Dabney.
\newblock Revisiting {{Fundamentals}} of {{Experience Replay}}.
\newblock 2020.

\bibitem[Fu et~al.(2022)Fu, Wu, and Boulet]{fu2022benchmarking}
Yuwei Fu, Di~Wu, and Benoit Boulet.
\newblock Benchmarking sample selection strategies for batch reinforcement
  learning, 2022.

\bibitem[Hessel et~al.(2018)Hessel, Modayil, Van~Hasselt, Schaul, Ostrovski,
  Dabney, Horgan, Piot, Azar, and Silver]{hessel2017rainbow}
Matteo Hessel, Joseph Modayil, Hado Van~Hasselt, Tom Schaul, Georg Ostrovski,
  Will Dabney, Dan Horgan, Bilal Piot, Mohammad Azar, and David Silver.
\newblock Rainbow: Combining improvements in deep reinforcement learning.
\newblock In \emph{Proceedings of the AAAI conference on artificial
  intelligence}, volume~32, 2018.

\bibitem[Hong et~al.(2023)Hong, Chen, Lin, Pajarinen, and
  Agrawal]{hong2023topological}
Zhang-Wei Hong, Tao Chen, Yen-Chen Lin, Joni Pajarinen, and Pulkit Agrawal.
\newblock Topological {{Experience Replay}}, 2023.

\bibitem[Horgan et~al.(2018)Horgan, Quan, Budden, Barth-Maron, Hessel,
  Van~Hasselt, and Silver]{horgan2018distributed}
Dan Horgan, John Quan, David Budden, Gabriel Barth-Maron, Matteo Hessel, Hado
  Van~Hasselt, and David Silver.
\newblock Distributed prioritized experience replay.
\newblock \emph{arXiv preprint arXiv:1803.00933}, 2018.

\bibitem[Igata et~al.(2021)Igata, Ikegaya, and Sasaki]{igata2021prioritized}
Hideyoshi Igata, Yuji Ikegaya, and Takuya Sasaki.
\newblock Prioritized experience replays on a hippocampal predictive map for
  learning.
\newblock 118\penalty0 (1):\penalty0 e2011266118, 2021.

\bibitem[Jaderberg et~al.(2016)Jaderberg, Mnih, Czarnecki, Schaul, Leibo,
  Silver, and Kavukcuoglu]{jaderberg2016reinforcement}
Max Jaderberg, Volodymyr Mnih, Wojciech~Marian Czarnecki, Tom Schaul, Joel~Z
  Leibo, David Silver, and Koray Kavukcuoglu.
\newblock Reinforcement learning with unsupervised auxiliary tasks.
\newblock \emph{arXiv preprint arXiv:1611.05397}, 2016.

\bibitem[Kingma \& Ba(2015)Kingma and Ba]{Kingma2015adam}
Diederik Kingma and Jimmy Ba.
\newblock Adam: A method for stochastic optimization.
\newblock In \emph{International Conference on Learning Representations}, San
  Diega, CA, USA, 2015.

\bibitem[Kobayashi(2024)]{kobayashi2024revisiting}
Taisuke Kobayashi.
\newblock Revisiting experience replayable conditions.
\newblock \emph{arXiv preprint arXiv:2402.10374}, 2024.

\bibitem[Kumar \& Nagaraj(2023)Kumar and Nagaraj]{kumar2023introspective}
Ramnath Kumar and Dheeraj Nagaraj.
\newblock Introspective {{Experience Replay}}: {{Look Back When Surprised}},
  2023.

\bibitem[Lee et~al.(2019)Lee, Sungik, and Chung]{lee2019sampleefficient}
Su~Young Lee, Choi Sungik, and Sae-Young Chung.
\newblock Sample-{{Efficient Deep Reinforcement Learning}} via {{Episodic
  Backward Update}}.
\newblock In \emph{Advances in {{Neural Information Processing Systems}}},
  volume~32. {Curran Associates, Inc.}, 2019.

\bibitem[Li et~al.(2021)Li, Lu, and Miao]{li2021revisiting}
Ang~A Li, Zongqing Lu, and Chenglin Miao.
\newblock Revisiting prioritized experience replay: A value perspective.
\newblock \emph{arXiv preprint arXiv:2102.03261}, 2021.

\bibitem[Li et~al.(2022)Li, Huang, and Zhu]{li2022clustering}
Min Li, Tianyi Huang, and William Zhu.
\newblock Clustering experience replay for the effective exploitation in
  reinforcement learning.
\newblock \emph{Pattern Recognition}, 131:\penalty0 108875, 2022.

\bibitem[Lin(1991)]{lin1991programming}
Long-Ji Lin.
\newblock Programming robots using reinforcement learning and teaching.
\newblock In \emph{Proceedings of the Ninth National Conference on Artificial
  Intelligence - Volume 2}, AAAI'91, pp.\  781–786. AAAI Press, 1991.

\bibitem[Lu et~al.(2024)Lu, Ball, Teh, and Parker-Holder]{lu2024synthetic}
Cong Lu, Philip Ball, Yee~Whye Teh, and Jack Parker-Holder.
\newblock Synthetic experience replay.
\newblock \emph{Advances in Neural Information Processing Systems}, 36, 2024.

\bibitem[Ma et~al.(2023)Ma, Li, Zhang, Liu, Wang, Chen, Shen, Wang, and
  Tao]{ma2023revisiting}
Guozheng Ma, Lu~Li, Sen Zhang, Zixuan Liu, Zhen Wang, Yixin Chen, Li~Shen,
  Xueqian Wang, and Dacheng Tao.
\newblock Revisiting {{Plasticity}} in {{Visual Reinforcement Learning}}:
  {{Data}}, {{Modules}} and {{Training Stages}}, 2023.

\bibitem[Mnih et~al.(2015)Mnih, Kavukcuoglu, Silver, Rusu, Veness, Bellemare,
  Graves, Riedmiller, Fidjeland, Ostrovski, et~al.]{mnih2015human}
Volodymyr Mnih, Koray Kavukcuoglu, David Silver, Andrei~A Rusu, Joel Veness,
  Marc~G Bellemare, Alex Graves, Martin Riedmiller, Andreas~K Fidjeland, Georg
  Ostrovski, et~al.
\newblock Human-level control through deep reinforcement learning.
\newblock \emph{nature}, 518\penalty0 (7540):\penalty0 529--533, 2015.

\bibitem[Moore \& Atkeson(1993)Moore and Atkeson]{moore1993prioritized}
Andrew~W. Moore and Christopher~G. Atkeson.
\newblock Prioritized {{Sweeping}}: {{Reinforcement Learning}} with {{Less
  Data}} and {{Less Time}}.
\newblock 13\penalty0 (1):\penalty0 103--130, 1993.
\newblock ISSN 0885-6125.

\bibitem[Moore(1990)]{moore1990efficient}
Andrew~William Moore.
\newblock Efficient memory-based learning for robot control.
\newblock Technical report, University of Cambridge, 1990.

\bibitem[Pan et~al.(2018)Pan, Zaheer, White, Patterson, and
  White]{pan2018organizing}
Yangchen Pan, Muhammad Zaheer, Adam White, Andrew Patterson, and Martha White.
\newblock Organizing {{Experience}}: {{A Deeper Look}} at {{Replay Mechanisms}}
  for {{Sample-based Planning}} in {{Continuous State Domains}}, 2018.

\bibitem[Patterson et~al.(2022)Patterson, White, and
  White]{patterson2022generalized}
Andrew Patterson, Adam White, and Martha White.
\newblock A generalized projected bellman error for off-policy value estimation
  in reinforcement learning.
\newblock 23\penalty0 (1):\penalty0 145:6463--145:6523, 2022.
\newblock ISSN 1532-4435.

\bibitem[Patterson et~al.(2023)Patterson, Neumann, White, and
  White]{patterson2023empirical}
Andrew Patterson, Samuel Neumann, Martha White, and Adam White.
\newblock Empirical {{Design}} in {{Reinforcement Learning}}, 2023.

\bibitem[Schaul et~al.(2015)Schaul, Horgan, Gregor, and
  Silver]{schaul2015universal}
Tom Schaul, Daniel Horgan, Karol Gregor, and David Silver.
\newblock Universal {{Value Function Approximators}}.
\newblock In \emph{Proceedings of the 32nd {{International Conference}} on
  {{Machine Learning}}}, pp.\  1312--1320. {PMLR}, 2015.

\bibitem[Schaul et~al.(2016)Schaul, Quan, Antonoglou, and
  Silver]{schaul2016prioritized}
Tom Schaul, John Quan, Ioannis Antonoglou, and David Silver.
\newblock Prioritized {{Experience Replay}}.
\newblock 2016.

\bibitem[Sun et~al.(2020)Sun, Zhou, and Li]{sun2020attentive}
Peiquan Sun, Wengang Zhou, and Houqiang Li.
\newblock Attentive {{Experience Replay}}.
\newblock 34\penalty0 (04):\penalty0 5900--5907, 2020.
\newblock ISSN 2374-3468.

\bibitem[Sutton et~al.(2009)Sutton, Maei, Precup, Bhatnagar, Silver,
  Szepesvári, and Wiewiora]{sutton2009fast}
Richard Sutton, Hamid Maei, Doina Precup, Shalabh Bhatnagar, David Silver,
  Csaba Szepesvári, and Eric Wiewiora.
\newblock Fast gradient-descent methods for temporal-difference learning with
  linear function approximation.
\newblock volume 382, pp.\  125, 2009.

\bibitem[Sutton(1988)]{sutton1988learning}
Richard~S Sutton.
\newblock Learning to predict by the methods of temporal differences.
\newblock \emph{Machine learning}, 3:\penalty0 9--44, 1988.

\bibitem[Sutton(1995)]{sutton1995generalization}
Richard~S Sutton.
\newblock Generalization in reinforcement learning: Successful examples using
  sparse coarse coding.
\newblock In D.~Touretzky, M.C. Mozer, and M.~Hasselmo (eds.), \emph{Advances
  in Neural Information Processing Systems}, volume~8. MIT Press, 1995.

\bibitem[Sutton \& Barto(2018)Sutton and Barto]{sutton2018reinforcement}
Richard~S Sutton and Andrew~G Barto.
\newblock \emph{Reinforcement learning: An introduction}.
\newblock MIT press, 2018.

\bibitem[Van~Hasselt et~al.(2019)Van~Hasselt, Hessel, and
  Aslanides]{van2019use}
Hado~P Van~Hasselt, Matteo Hessel, and John Aslanides.
\newblock When to use parametric models in reinforcement learning?
\newblock \emph{Advances in Neural Information Processing Systems}, 32, 2019.

\bibitem[Wang et~al.(2024)Wang, Miahi, White, Machado, Abbas, Kumaraswamy, Liu,
  and White]{wang2024investigating}
Han Wang, Erfan Miahi, Martha White, Marlos~C Machado, Zaheer Abbas, Raksha
  Kumaraswamy, Vincent Liu, and Adam White.
\newblock Investigating the properties of neural network representations in
  reinforcement learning.
\newblock \emph{Artificial Intelligence}, pp.\  104100, 2024.

\bibitem[Wittkuhn et~al.(2021)Wittkuhn, Chien, Hall-McMaster, and
  Schuck]{wittkuhn2021replay}
Lennart Wittkuhn, Samson Chien, Sam Hall-McMaster, and Nicolas~W. Schuck.
\newblock Replay in minds and machines.
\newblock 129:\penalty0 367--388, 2021.
\newblock ISSN 0149-7634.

\end{thebibliography}
\bibliographystyle{rlc}

\appendix

\section{Algorithms}

Here we present pseudo code for the DQN algorithm used in this paper. The behavior policy of DQN, denoted as $\pi_{\bf w}(S_t)$, is an $\epsilon$-greedy policy over the current action values $q(S_t,.)$. Algorithm \ref{alg:dqn} shows the uniform replay variant and algorithm \ref{alg:naive-per} shows the pseudo code for \pername. In both variants, size of the replay buffer is a hyperparameter.

\begin{algorithm}[htb]
    \caption{DQN with Uniform Replay}\label{alg:dqn}
 \begin{algorithmic}
    \STATE {\bfseries Input:} mini-batch size $b$, learning-rate $\alpha$, training time $T$, target refresh rate $\tau$.
    \STATE {\bfseries Initialize:} $q$ network parameters ${\bf w}$, target network parameters ${\bf w}_\text{target} = {\bf w}$, buffer $B$, $\Delta = 0$.
    \STATE Observe $S_0$ and choose $A_0 \sim \pi_{\bf w}(S_0)$
    \FOR{$t=1$ {\bfseries to} $T$} 
        \STATE Observe $R_t, S_t, \gamma_t$
        \STATE Store transition $(S_{t-1}, A_{t-1}, R_t, S_{t}, \gamma_{t})$ in buffer $B$
        \FOR{$j=1$ {\bfseries to} $b$}
            \STATE Sample transition $(S_j, A_j, R_j, S_{j+1}, \gamma_{j+1})$ from buffer $B$ with probability $1/|B|$
            \STATE Compute TD-error $\delta_j =  R_j + \gamma_j \max_a q(S_{j+1}, a, {\bf w}_{\text{target}}) - q(S_j, A_j, {\bf w})$
            \STATE Accumulate gradient $\Delta \leftarrow \Delta - \delta_j \nabla_{\bf w}q(S_j, A_j, {\bf w})$
        \ENDFOR
        \STATE Update ${\bf w} \leftarrow \text{adam}({\bf w}, \frac{\Delta}{b}, \alpha)$; Reset $\Delta = 0$
        \IF{$t \% \tau = 0$}
            \STATE Refresh target network ${\bf w}_{\text{target}} \leftarrow {\bf w}$
        \ENDIF
        \STATE Choose action $A_t \sim \pi_{\bf w}(S_t)$
    \ENDFOR
    \end{algorithmic}
\end{algorithm}

\begin{algorithm}[htb]
    \caption{DQN with \pername}
    \label{alg:naive-per}
 \begin{algorithmic}
    \STATE {\bfseries Input:} mini-batch size $b$, learning-rate $\alpha$, training time $T$, target refresh rate $\tau$.
    \STATE {\bfseries Initialize:} $q$ network parameters ${\bf w}$, target network parameters ${\bf w}_\text{target} = {\bf w}$, prioritized buffer $B$, $\Delta = 0$.
    \STATE Observe $S_0$ and choose $A_0 \sim \pi_{\bf w}(S_0)$
    \FOR{$t=1$ {\bfseries to} $T$} 
        \STATE Observe $R_t, S_t, \gamma_t$
        \STATE Store transition $(S_{t-1}, A_{t-1}, R_t, S_t, \gamma_{t})$ in buffer $B$ with priority $p_{t-1} = |\delta_{t-1}| = |R_t + \gamma_t \max_a q(S_t, a, {\bf w}_{\text{target}}) - q(S_{t-1}, A_{t-1}, {\bf w})|$
        \FOR{$j=1$ {\bfseries to} $b$}
            \STATE Sample transition $(S_j, A_j, R_j, S_{j+1}, \gamma_{j+1})$ from buffer $B$ with probability $\frac{p_j}{\sum_ip_i}$
            \STATE Compute TD-error $\delta_j =  R_j + \gamma_j \max_a q(S_{j+1}, a, {\bf w}_{\text{target}}) - q(S_j, A_j, {\bf w})$
            \STATE Accumulate gradient $\Delta \leftarrow \Delta - \delta_j \nabla_{\bf w}q(S_j, A_j, {\bf w})$
        \ENDFOR
        \FOR{$j=1$ {\bfseries to} $b$}
            \STATE Update priority $p_j = |\delta_j|$
        \ENDFOR
        \STATE Update ${\bf w} \leftarrow \text{adam}({\bf w}, \frac{\Delta}{b}, \alpha)$; Reset $\Delta = 0$
        \IF{$t \% \tau = 0$}
            \STATE Refresh target network ${\bf w}_{\text{target}} \leftarrow {\bf w}$
        \ENDIF
        \STATE Choose action $A_t \sim \pi_{\bf w}(S_t)$
    \ENDFOR
    \end{algorithmic}
\end{algorithm}

\section{DM-PER Hyperparameters}

In all the experiments with DM-PER, its additional hyperparameters are set to the values from \citep{schaul2016prioritized} and are listed in table \ref{table:dm-per}.

\begin{table}[htb]
\begin{center}
\begin{tabular}{ccc} \hline
Priority exponent & Importance sampling exponent & i.i.d mix-in ratio \\ \hline
$0.6$ & $0.4\rightarrow1.0$ & $10^{-3}$\\
\hline
\end{tabular}
\caption{Hyperparameters specific to DM-PER. Arrow indicates linear schedule over training time.}
\label{table:dm-per}
\end{center}
\end{table}

\section{Prediction Experiments in the Chain}

In This section, we document the hyperparameter sweep details and additional results from the Markov chain prediction experiment. Table \ref{table:pred} lists the hyperparameter selections for the prediction agents. The agents in figure \ref{fig:chain-pred} use buffer size $8000$, batch size $8$, and learning rate $8^{-4}$ for tabular and $8^{-5}$ for neural network agents.

\begin{table}[htb]
\begin{center}
\begin{tabular}{ccc} \hline
& Tabular agents & Neural network agents \\ \hline
Learning rate & $[8^{-6}, 8^{-5}, 8^{-4}, 8^{-3}, 8^{-2}]$ & $[8^{-6}, 8^{-5}/4, 8^{-5}, 8^{-4}/4, 8^{-4}, 8^{-3}/4, 8^{-3}]$\\
Adam optimizer $\beta_1$ & $0.9$ & $0.9$\\
Adam optimizer $\beta_2$ & $0.999$ & $0.999$\\
Batch size & $[1, 8, 64]$ & $[1, 8, 64]$\\
Buffer size & $[800, 8000, 80000]$ & $[800, 8000, 80000]$\\
Network size & -  & $2\times32$ network with ReLU activation\\
Training time & $80000$ & $80000$\\
\hline
\end{tabular}
\caption{Hyperparameters of prediction agents in Markov chain}
\label{table:pred}
\end{center}
\end{table}

\begin{figure}[htb]
\vspace{-0.2cm}
\begin{center}
\includegraphics[width=1.0\columnwidth]{"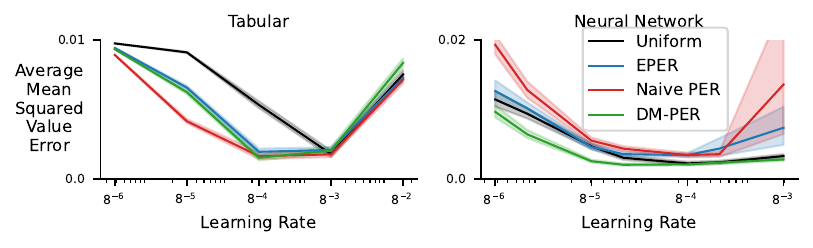"}
\vspace{-0.5cm}
\caption{Sensitivity to learning rate in prediction chain task. Results are averaged over 30 seeds and shaded region is 95\% bootstrap CI.}
\label{fig:pred-sensitivity}
\end{center}
\vspace{-0.2cm}
\end{figure}

 Figure \ref{fig:pred-sensitivity} shows the sensitivity of replay methods to learning rate for batch size 8 and buffer size 8000 in the chain prediction problem. Prioritized replay is more sample efficient than uniform replay in the tabular setting, especially with smaller step sizes. But when using neural networks, the early increase in MSVE of \pername, reduces its average performance below other algorithms.

\begin{figure}[htb]
\vspace{-0.2cm}
\begin{center}
\includegraphics[width=1.0\columnwidth]{"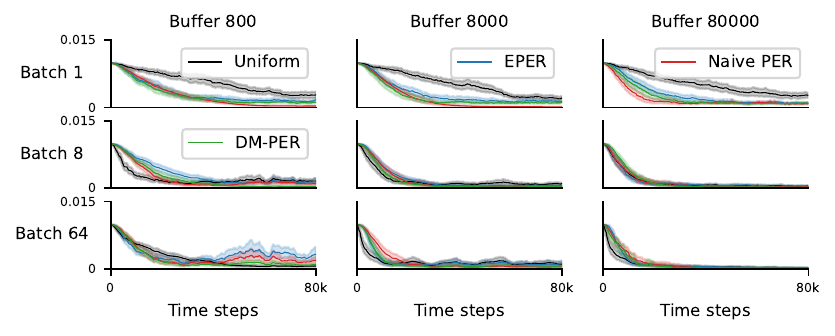"}
\vspace{-0.2cm}
\caption{Performance of tabular replay agents in the prediction chain task. Results are averaged over 30 seeds; shaded region is 95\% bootstrap CI.}
\label{fig:pred-tabular-more}
\end{center}
\vspace{-0.2cm}
\end{figure}

\begin{figure}[htb]
\vspace{-0.2cm}
\begin{center}
\includegraphics[width=1.0\columnwidth]{"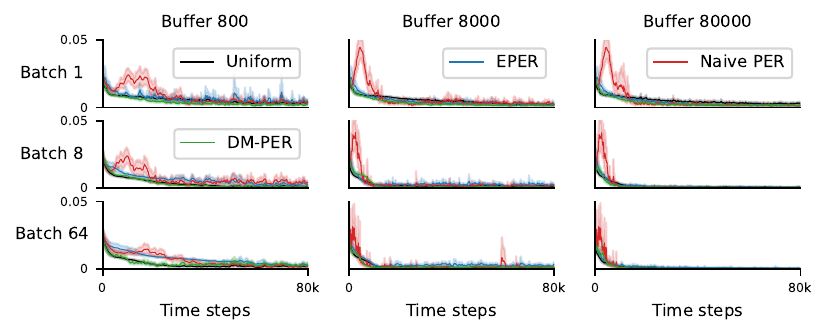"}
\vspace{-0.2cm}
\caption{Performance of neural network replay agents in the prediction chain task. Results are averaged over 30 seeds; shaded region is 95\% bootstrap CI.}
\label{fig:pred-nn-more}
\end{center}
\vspace{-0.2cm}
\end{figure}

Now we show additional results for those meta-parameter choices in the chain prediction domain that are omitted from the main text. Figures \ref{fig:pred-tabular-more} and \ref{fig:pred-nn-more} show the learning curves for tabular and neural network agents respectively. The learning rate is tuned by maximizing over average performance across 30 seeds.

\begin{figure*}[htb]
\vspace{-0.2cm}
\begin{center}
\includegraphics[width=1.0\columnwidth]{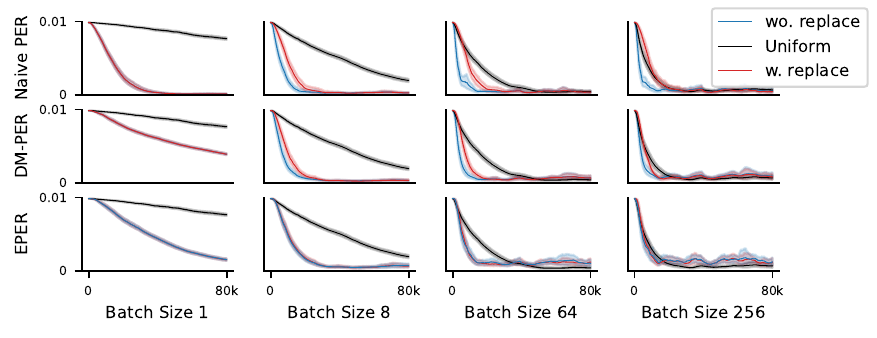}
\vspace{-0.5cm}
\caption{Sampling without replacement improves performance in the tabular prediction chain problem for \pername\ and DM-PER. Results averaged over 50 seeds with 95\% bootstrap CI.}
\label{fig:pred-tabular-replacement-more}
\end{center}
\vspace{-0.2cm}
\end{figure*}

\begin{figure*}[htb]
\vspace{-0.2cm}
\begin{center}
\includegraphics[width=1.0\columnwidth]{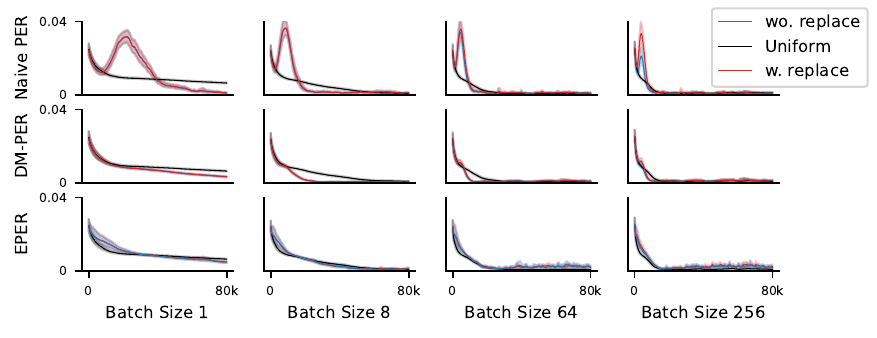}
\vspace{-0.5cm}
\caption{Sampling without replacement does not improve performance in the prediction chain problem under neural network function approximation. Results averaged over 50 seeds with 95\% bootstrap CI.}
\label{fig:pred-nn-replacement-more}
\end{center}
\vspace{-0.2cm}
\end{figure*}

As part of our investigation into PER, we experimented with sampling mini-batches without replacement (see figure \ref{fig:pred-replacement}). Here we present results for all PER variants, \pername, DM-PER, and EPER in the aforementioned prediction experiment. The hyperparameters of this experiment are chosen according to table \ref{table:pred} sweeping over a range of batch sizes $[1, 8, 64, 256]$. We chose the same learning rate as in figure \ref{fig:chain-pred}, namely, $8^{-4}$ for tabular agents and $8^{-5}$ for neural network agents. Figure \ref{fig:pred-tabular-replacement-more} shows tabular agents and figure \ref{fig:pred-nn-replacement-more} shows the neural network agents MSVE over training time.


\section{Control Markov Chain Experiment Details}

The hyperparameters used in control chain experiments are given in table \ref{table:control-chain}. We chose the learning rate for each agent by maximizing over average performance across a range of learning rates.

\begin{table}[htb]
\begin{center}
\begin{tabular}{ccc} \hline
& Q-Learning agents (tabular) & DQN and EQRC agents (neural network) \\ \hline
Learning rate & $[8^{-7}, 8^{-6}, 8^{-5}, 8^{-4}, 8^{-3}, 8^{-2}]$ & $[8^{-5}, 8^{-4}, 8^{-3}, 8^{-2}, 8^{-1}]$\\
Adam optimizer $\beta_1$ & $0.9$ & $0.9$\\
Adam optimizer $\beta_2$ & $0.999$ & $0.999$\\
Batch size & $8$ & $8$\\
Buffer size & $10000$ & $10000$\\
Network size & -  & $2\times32$ network with ReLU activation\\
Target refresh & - & 100 (only DQN)\\
Exploration $\epsilon$ & 0.1 & 0.1\\
Training time & $100000$ & $100000$\\
\hline
\end{tabular}
\caption{Hyperparameters of control agents in Markov chain}
\label{table:control-chain}
\end{center}
\end{table}

\begin{figure*}[htb]
\vspace{-0.2cm}
\begin{center}
\includegraphics[width=1.0\columnwidth]{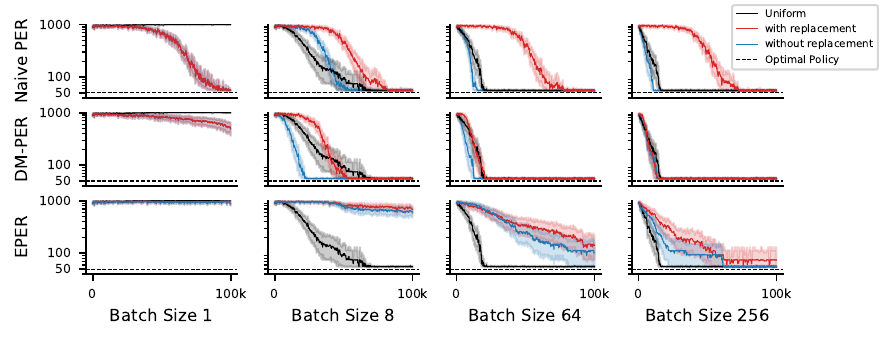}
\vspace{-0.5cm}
\caption{Sampling without replacement improves performance in the control chain task when using tabular Q-learning. Results are averaged over 50 seeds; shaded region is 95\% boostrap CI.}
\label{fig:control-q-replacement-more}
\end{center}
\vspace{-0.2cm}
\end{figure*}

\begin{figure*}[htb]
\vspace{-0.2cm}
\begin{center}
\includegraphics[width=1.0\columnwidth]{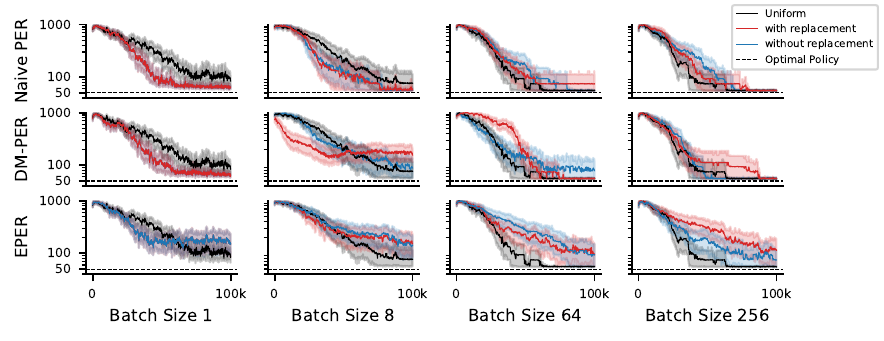}
\vspace{-0.5cm}
\caption{Sampling without replacement does not improve performance in control chain task when using DQN. Results are averaged over 50 seeds; shaded region is 95\% boostrap CI.}
\label{fig:control-dqn-replacement-more}
\end{center}
\vspace{-0.2cm}
\end{figure*}

\begin{figure*}[htb]
\vspace{-0.2cm}
\begin{center}
\includegraphics[width=1.0\columnwidth]{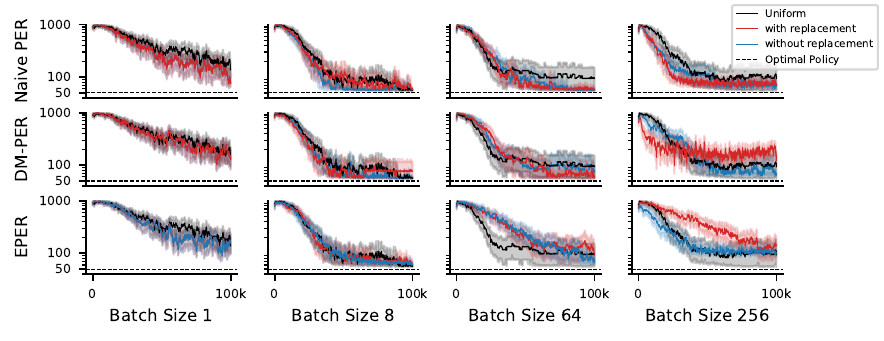}
\vspace{-0.5cm}
\caption{Sampling without replacement does not improve performance in the control chain task when using EQRC. Results are averaged over 50 seeds; shaded region is 95\% boostrap CI.}
\label{fig:control-eqrc-replacement-more}
\end{center}
\vspace{-0.2cm}
\end{figure*}

In the without replacement control experiment, we use a buffer size of 10000 and experiment with 4 different batch sizes $[1, 8, 64, 256]$. For each setting, the learning-rate is selected via maximizing over a range of learning-rates. All other meta-parameters are based on table \ref{table:control-chain}. Figure \ref{fig:control-q-replacement-more} shows results for the tabular Q-learning setting, figure \ref{fig:control-dqn-replacement-more} shows the DQN results, and figure \ref{fig:control-eqrc-replacement-more} shows the EQRC results.

\section{Classic Control Experiment Details}

\subsection{Environment Description}

In the last part of the paper, we experiment with classic control problems that are more difficult than the chain. We experiment with MountainCar, Acrobot, Cartpole, and Cliffworld. In MountainCar, the goal is to drive an under powered car up a hill in a simulated environment with simplified physics by taking one of three actions, accelerate left, accelerate right, do not accelerate. The observations are position and speed values of the car. The reward is -1 per step and episodes terminates when the car crosses a threshold at the top of hill with reward 0.

In Acrobot the agent controls a system of two linear links connected by a movable joint. The goal is to move the links, by applying torque to the joint, such that the bottom part of the link rises to the level of its highest point upon which the episode terminates with reward 0. The reward of all other transitions is -1 per step.

The goal of a Cartpole agent is to balance a pole on top of a moving cart by accelerating the cart to either left or right. The reward is +1 per step if the pole is properly balanced. If the pole falls more than 12 degrees the episodes is terminated and the pole is reset to its upright position. The episode cutoff length is 500.

Cliffworld is a gridworld where agents start at a fixed state and pick any of cardinal directions and move to corresponding neighbor state. The goal is to reach to the final state on the opposite side of the starting state while avoiding a cliff near the optimal path. The reward is -1 per step except when falling off the cliff that gives -100 reward upon which the agent is reset back to start (without episode termination).

\subsection{Hyperparameter selection}

In the classic control experiments we use the hyperparameters from table \ref{table:cc} and tune the step size using the two stage hyperparameter selection method \citep{patterson2023empirical}. For each agent we run all learning rates for 30 seeds, selecting the value with maximum average performance, then running the tuned agent for 100 new seeds to avoid maximization bias.

\begin{table}[htb]
\begin{center}
\begin{tabular}{cc} \hline
& DQN agents \\ \hline
Learning rate & $[4^{-8}, 4^{-7}, 4^{-6}, 4^{-5}, 4^{-4}, 4^{-3}, 4^{-2}]$\\
Adam optimizer $\beta_1$ & $0.9$\\
Adam optimizer $\beta_2$ & $0.999$\\
Batch size & $64$\\
Buffer size & $10000$\\
Network size &$2\times64$ dense network with ReLU activation\\
Target refresh & 128\\
Exploration $\epsilon$ & 0.1\\
Training time & $100000$\\
\hline
\end{tabular}
\caption{Hyperparameters of classic control experiments}
\label{table:cc}
\end{center}
\end{table}

\end{document}